\documentclass[conference]{IEEEtran}
\usepackage{cite}
\usepackage{amsmath,amssymb,amsfonts}
\usepackage{algorithmic}
\usepackage{graphicx}
\usepackage{textcomp}
\usepackage{xcolor}

\usepackage{multirow}
\usepackage[caption=false]{subfig}

\def\BibTeX{{\rm B\kern-.05em{\sc i\kern-.025em b}\kern-.08em
    T\kern-.1667em\lower.7ex\hbox{E}\kern-.125emX}}
\begin{document}

\title{E2EG: End-to-End Node Classification Using Graph Topology and Text-based Node Attributes
}

\author{\IEEEauthorblockN{1\textsuperscript{st} Tu Anh Dinh}
\IEEEauthorblockA{\textit{Karlsruhe Institute of Technology}\\
Karlsruhe, Germany \\
0000-0001-7651-820X}
\and
\IEEEauthorblockN{2\textsuperscript{nd} Jeroen den Boef}
\IEEEauthorblockA{\textit{Helin Data}\\
Amsterdam, The Netherlands \\
0000-0001-5649-2778}
\and
\IEEEauthorblockN{3\textsuperscript{rd} Joran Cornelisse}
\IEEEauthorblockA{\textit{SocialDatabase}\\
Amsterdam, The Netherlands \\
0000-0002-1244-9900}
\and
\IEEEauthorblockN{4\textsuperscript{th} Paul Groth}
\IEEEauthorblockA{\textit{University of Amsterdam}\\
Amsterdam, The Netherlands \\
0000-0003-0183-6910}
}

\maketitle

\begin{abstract}
Node classification utilizing text-based node attributes has many real-world applications, ranging from prediction of paper topics in academic citation graphs to classification of user characteristics in social media networks. State-of-the-art node classification frameworks, such as GIANT, use a two-stage pipeline: first embedding the text attributes of graph nodes then feeding the resulting embeddings into a node classification model. In this paper, we eliminate these two stages and develop an {\em end-to-end} node classification model that builds upon GIANT, called End-to-End-GIANT (E2EG). The tandem utilization of a main and an auxiliary classification objectives in our approach results in a more robust model, enabling the BERT backbone to be switched out for a distilled encoder with a 25\% - 40\% reduction in the number of parameters. Moreover, the model's end-to-end nature increases ease of use, as it avoids the need of chaining multiple models for node classification. Compared to a GIANT+MLP baseline on the \textit{ogbn-arxiv} and \textit{ogbn-products} datasets, E2EG obtains slightly better accuracy in the transductive setting (+0.5\%), while reducing model training time by up to 40\%. Our model is also applicable in the inductive setting, outperforming GIANT+MLP by up to +2.23\%.
\end{abstract}


\section{Introduction}
Recently, machine learning on graphs has gained more attention due to its real-world applicability \cite{KipfWelling,GraphSAGE,GAT,GraphSAINT,GIANT}. Classifying papers' topics in citation networks \cite{MAG_citation_net}, predicting links between users in social networks \cite{linkpred_socialnet} or classifying the types of molecular structures \cite{graphpred_molecular} can all be represented as graph-based problems. One group of graph-based problems is node classification, where the aim is to predict a characteristic of the entities (nodes) belonging to a graph given their attributes. In many cases, the nodes' attributes are text, for instance, social media user bios or scientific paper abstracts \cite{social_net,MAG_citation_net,AmazonProducts,NetData}. From this point onward, we refer to node classification with text-based node attributes as \textit{textual node classification}, which is the focus of this paper.

To perform textual node classification, it is useful to leverage all available modalities: the nodes' text attributes and the graph topology. Textual node classification often includes two stages: 1) embedding the nodes' raw text attributes; and 2) using the embeddings as input to a downstream classifier. A common approach is to use graph-agnostic text embeddings such as bag-of-words or word2vec as input to graph-dependent classifiers such as Graph Neural Networks (GNNs) \cite{KipfWelling,GraphSAGE,GAT,GraphSAINT}. 

The current state-of-the-art framework, GIANT \cite{GIANT}, further exploits the correlation between graph topology and raw text by generating graph-dependent text embeddings to be used in downstream node classification models. GIANT embeds the nodes' raw text by training an encoder on a self-supervised learning task termed "neighborhood prediction". The task's objective is to predict the neighborhood of a node given the node's raw text attributes, thus incorporating graph topology while embedding the raw text. The inclusion of GIANT's graph-dependent text embeddings results in improved performance for graph-agnostic downstream classifiers (e.g., a multilayer perceptron (MLP)), thus reducing the need for computationally expensive GNNs.

In GIANT and other two-stage frameworks, information loss from text attributes could occur due to objective switching between the self-supervised text embedding process and the subsequent node classification process. Here, the nodes’ raw text is not embedded directly under the main node classification objective, thus could have a negative effect on node classification quality. 
This could potentially be addressed by simultaneously utilizing graph topology and embedding the nodes' raw text for node classification in an end-to-end fashion, which has not yet been explored. Additionally, employing both raw text and graph structure in tandem for training could result in a more robust model due to increased objective expressivity, thus allowing the usage of a more lightweight text encoder with improved computational efficiency.

\textbf{Present work.} With the aim of preventing information loss and reducing computational cost, we propose an end-to-end model that utilizes both graph topology and nodes' raw text directly for node classification, called E2EG (End-to-end GIANT) \footnote{Code available at https://github.com/TuAnh23/E2EG.}. 
We make use of the neighborhood prediction task in the GIANT framework \cite{GIANT}. Neighborhood prediction has shown to be an effective task to include graph topology implicitly, as opposed to computationally expensive GNNs that explicitly use feature propagation operations over neighboring nodes \cite{GraphSAGE,GAT,GraphSAINT}. Our E2EG model takes the node's raw text attributes as input, and learns both the main node classification task and the neighborhood prediction task in tandem. In this way, E2EG simultaneously embeds the text under the main objective, while utilizing graph topology via neighborhood prediction. 

In summary, the contributions of this paper are as follows: 
\begin{enumerate}
\item A new end-to-end model for node classification, E2EG, that is well-suited for real-world usage due to its compact and stand-alone nature which excludes the need to chain multiple models for node classification;
\item Experiments showing that, in contrast to GIANT, E2EG is able to use a lightweight and faster distilled text encoder without accuracy loss;
\item Experiments showing that E2EG outperforms GIANT+MLP in both transductive and inductive settings.
\item Qualitative analysis indicating that E2EG: 1) potentially reduces information loss from text attributes compared to the two-staged GIANT+MLP; and 2) better exploits topological information compared to a BERT text classification model thanks to the multi-task setting.
\end{enumerate}

\section{Background and related work} \label{sec:BackgroundRelatedWork}
In this section, we give a formal definition of a graph, followed by the common approaches for general machine learning on graphs and the computational overhead inherent to these frameworks. Afterward, we describe the related works on textual node classification. Finally, we give a detailed background description of the GIANT framework for textual node classification, on which we base our E2EG model.

\subsection{Graph}
A graph is a structure used to represent entities that are related to each other. A graph $G$ consists of a node set $\mathcal{V} = \{v_1,v_2...,v_n\}$, and an edge set which can be represented by an adjacency matrix $A \in \{0,1\}^{n \times n}$. Each node $v_i$ can have attributes denoted as $T_i$. A slice $A_i$ of the adjacency matrix denotes the one-hot-encoded neighborhood $N(v_i)$ of node $v_i$.

\subsection{Machine learning on graphs} 
A common approach to machine learning on graphs, including for node classification, is through message-passing Graph Neural Networks (GNNs) \cite{GraphSAGE,GAT,GraphSAINT}. For example,  GraphSAGE \cite{GraphSAGE} updates the representation of each node $v_i$ by aggregating the representations of its direct neighbors $N(v_i)$ and concatenating it with the current representation of $v_i$. While proving to be effective, these message-passing GNNs can be computationally expensive, as message passing involves a growing number of support nodes required to calculate the representation of one node as the model goes deeper. This bottleneck is more widely known as the "neighbor explosion" problem \cite{GraphSAINT}.

\subsection{Textual node classification}
Textual node classification frameworks often embed the text to be used in downstream classifiers \cite{VectorizeTextForNodeClassification1,VectorizeTextForNodeClassification2,VectorizeTextForNodeClassification3,VectorizeTextForNodeClassification4}. Conventionally, graph-agnostic text embeddings are used as input to Graph Neural Network (GNN) classifiers in order to utilize the textual and graph modalities \cite{KipfWelling,GraphSAGE,GAT,GraphSAINT}. Examples of graph-agnostic text embeddings include bag-of-words features followed by a Principal Component Analysis; average of word vectors generated by a word2vec model \cite{word2vec}.

More recent attempts at graph-agnostic text embedding favor transformer-based approaches. Transformers \cite{transformer} consider the relative position of words within a sentence, hence the resulting embeddings can be more expressive compared to bag-of-words and word2vec. Additionally, transformers can be used directly as text classification models \cite{BERTClassifier} to classify the nodes, avoiding the need for a downstream node classifier and increasing ease of use. In this way, the model predicts the nodes' classes given the nodes' raw text attributes, but the graph topology is not utilized. 

Several works have proposed methods to incorporate both text data and graph topology to generate node embeddings, which can later be used for node classification. Paper2vec \cite{paper2vec} learns node embeddings by training a model to identify whether two nodes are from the same neighborhood given their text attributes. The authors in \cite{LinkPrediction} use link prediction given the nodes' raw text to learn node embeddings. Heterformer \cite{Heterformer} stacks graph-attention-based and transformer-based modules to embed nodes. In these papers, the resulting node embeddings are used as input for graph-agnostic node classifiers, e.g., Support Vector Machines, Multilayer Perceptron or Logistic Regression.

\subsection{The GIANT framework}

\begin{figure*}[!ht]
  \centering
  \includegraphics[width=0.95\textwidth]{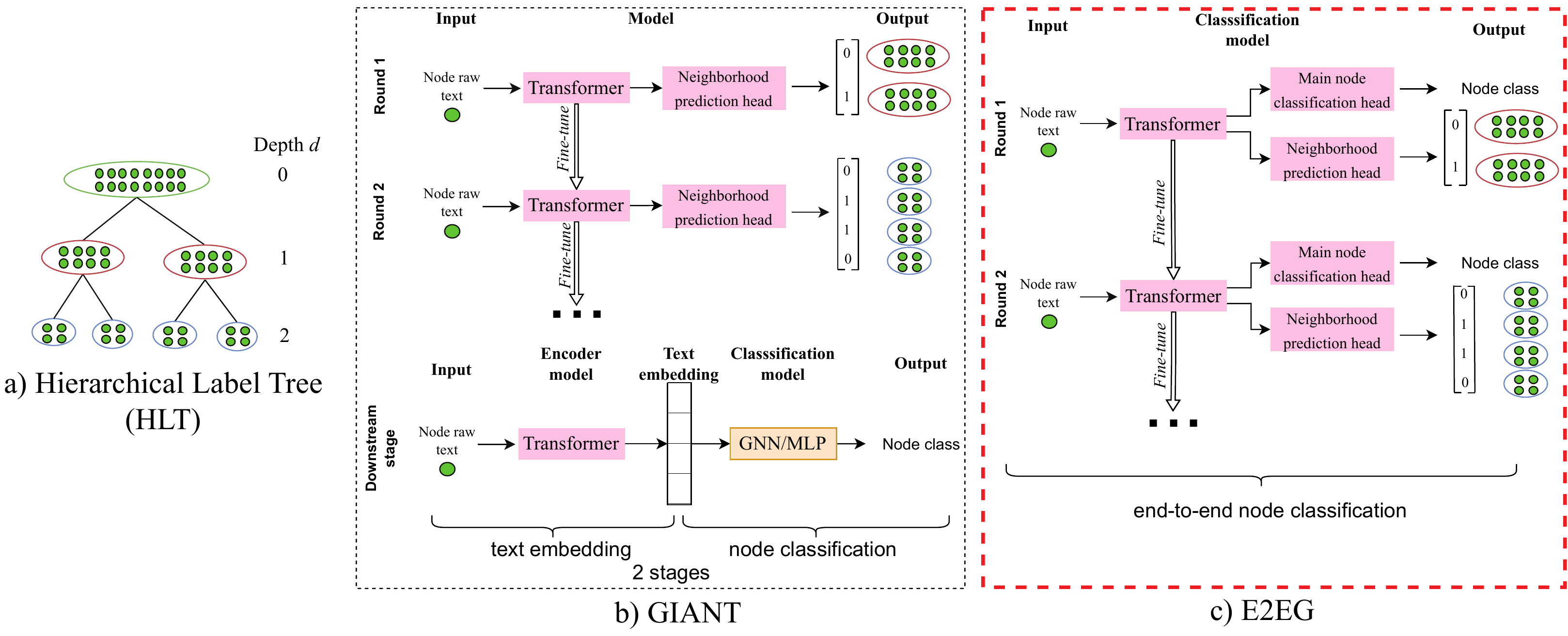}
  \caption{Illustrations of GIANT and the proposed E2EG model. a) Hierarchical Label Tree (HLT): the node set is recursively clustered to be used as targets for neighborhood prediction. b) GIANT recursively fine-tunes a transformer under the coarse-to-detailed neighborhood prediction objectives, then uses the resulting transformer to encode the nodes' raw text for a downstream node classifier. c) E2EG recursively fine-tunes a transformer model under both the neighborhood prediction and the main node classification objectives in an end-to-end fashion.}
  \label{fig:E2EGvsGIANT}
\end{figure*}

The GIANT framework \cite{GIANT} is currently one of the best in many OGB node classification leaderboards \cite{OGB}. GIANT generates graph-dependent text embeddings for downstream classifiers by employing a self-supervised learning task termed "neighborhood prediction". Consider a graph $G$ where each node $v_i$ is associated with some raw text $T_i$. The neighborhood prediction task's objective is to predict the one-hot-encoded neighborhood $A_i$ given $T_i$. By learning an encoder for nodes' raw text in order to predict nodes' neighborhoods, GIANT exploits the correlation between neighborhood topologies and their associated texts, thus utilizing graph topology to enhance text embeddings.

In neighborhood prediction, the output space is large since the size of the target vector $A_i$ is equal to the number of nodes $n$ in the graph. Neighborhood prediction is then an instance of the eXtreme Multi-label Classification (XMC) problem \cite{GIANT}. Therefore, GIANT uses the XR-Transformer architecture \cite{XR-transformer}, specifically designed for XMC. Adhering to XR-Transformer, the node set is first recursively clustered into a hierarchical label tree (HLT) based on the nodes' TF-IDF text features (Figure \ref{fig:E2EGvsGIANT}a). A transformer model is then recursively fine-tuned for neighborhood prediction, as illustrated in Figure \ref{fig:E2EGvsGIANT}b. At each fine-tuning round $d^{th}$, the transformer learns to map the nodes' text $T_i$'s to their clusters of neighbors defined by the HLT at depth $d$. The clusters' sizes get smaller after each fine-tuning round, meaning that the neighborhood to be predicted gets more detailed. This incrementally increased difficulty of the neighborhood prediction is a type of curriculum learning, where the weights of the $d^{th}$ transformer trained on the easier task (predicting coarser neighborhoods) are used to initialize the subsequent $(d+1)^{th}$ transformer trained on the more difficult task (predicting more detailed neighborhoods). GIANT then uses the text embeddings generated by the last fine-tuned transformer as input for a downstream node classifier.

The use of GIANT's graph-dependent text embeddings results in performance improvements for downstream node classifiers. The most remarkable improvement occurs for a graph-agnostic multilayer perceptron (MLP) classifier, shrinking the performance gap between this simple MLP and computationally expensive GNNs. In other words, GIANT+MLP obtains close performance to GIANT+GNN. 

The previous textual node classification frameworks mentioned above follow the common two-stage setup: first embedding the nodes' text attributes, then feeding the resulting embeddings as input to a downstream node classifier. Information loss could occur due to nodes' text attributes not being embedded directly for the main objective. To the best of our knowledge, the possibility of embedding the graph topology and the nodes' raw text directly for node classification in an end-to-end fashion has not yet been explored. With the aim of avoiding information loss and improving node classification quality in terms of accuracy and computational efficiency, we explore this direction by proposing an end-to-end node classification model, termed End-to-end GIANT (E2EG). The model is described in the following section.

\section{Methodology} \label{sec:Methodology}
We propose E2EG - a multi-task model that simultaneously utilizes graph topology and embeds nodes' raw text directly for node classification in an end-to-end fashion. E2EG's architecture is shown in Figure \ref{fig:E2EGvsGIANT}c. The model learns two tasks:
\begin{itemize}
    \item The main node classification task.\\
    Given input nodes' raw text, the model learns to predict the desired node classes. In this way, the raw text is directly embedded under the node classification objective.
    \item The auxiliary neighborhood prediction task.\\
    Inspired by the effectiveness of GIANT, we use neighborhood prediction as an auxiliary task for E2EG. In this way, the graph topology is encoded simultaneously with the text attributes, reducing the need for computationally expensive GNNs. Additionally, we hypothesize that the expressive learning objective due to the inclusion of an auxiliary task could allow for the usage of a more lightweight text encoder.
\end{itemize}

E2EG is expected to reduce information loss from text attributes (an issue hypothetically inherent to two-stage frameworks) due to its end-to-end nature, i.e., embedding nodes’ raw text directly for node classification. The multi-task setting facilitates this end-to-end training by making use of both raw text (via main node classification task) and the graph topology (via neighborhood prediction task) simultaneously. 

Since we use the same neighborhood prediction task as GIANT, we also build upon XR-Transformer \cite{XR-transformer}. E2EG takes a node's raw text as input, and outputs the node's class along with the node's one-hot-encoded neighborhood. Thus we modify XR-Transformer by adding a prediction head for the main node classification task. Our extended XR-Transformer includes 3 parts: the transformer component (i.e., the text encoder), the main classification head and the neighborhood prediction head (Figure \ref{fig:E2EGvsGIANT}c). Adhering to XR-Transformer, E2EG is trained over multiple fine-tuning rounds where the neighborhood prediction targets get more detailed. The main node classification task is fixed over fine-tuning rounds. The losses for neighborhood prediction and node classification are backpropagated in tandem.

We propose methods to improve the base E2EG model in terms of accuracy and computational efficiency as follows.

\subsection{Main task delay} \label{sec:main_task_delay}
Training the same node classification task along with different resolutions of neighborhood prediction over multiple fine-tuning rounds could lead to overfitting on the node classification task. The node classification performance could peak at an early fine-tuning round, hence not making use of the more detailed neighborhood information in the later rounds. Therefore, we propose delaying learning the main node classification task, by excluding the main task loss and only learning neighborhood prediction in earlier fine-tuning rounds.

\subsection{Additional fine-tuning round}
To optimize the node classification performance, we include an additional fine-tuning round in the end where the model only learns the main node classification task, and excludes the auxiliary neighborhood prediction task. In order to preserve the neighborhood information learnt in the previous rounds, we freeze the transformer component in the additional round, and only fine-tune the node classification head.

\subsection{Transductive data utilization}
In the transductive setting, the entire graph structure is available during training by definition. Therefore, it would be useful to exploit the test and validation nodes' topology for training. We do so by backpropagating only the neighborhood prediction loss and excluding the main task loss for the test and validation samples. In this way, we fully utilize the data while making sure that there is no leakage of the actual test and validation target classes. This is analogous to the data usage of GIANT, since GIANT's model is trained on the whole graph for neighborhood prediction, i.e., including all nodes in the train, validation and test sets.

\subsection{Encoders}
The expressive learning objective and end-to-end nature of our model could potentially allow for the employment of a more lightweight and faster transformer text encoder. We propose using distilled versions of the transformer to replace the original BERT component \cite{BERT} used by GIANT. The first version is DistilBERT \cite{DistilBERT}. We select DistilBERT as it would be a fair comparison to the BERT encoder used by GIANT. The second version is Distil-SentenceBERT \cite{sbert}.
We select this encoder as its pretraining objective is on the sentence level, rather than the word/token level. This could be more suitable for the task at hand, since for GIANT and E2EG, the objective is on the sentence level, i.e., embedding a piece of text or classifying a node based on a piece of text.

\section{Experimental Setup} \label{sec:ExperimentalSetup}
We train and evaluate our models on the \textit{ogbn-arxiv} \cite{MAG_citation_net} and \textit{ogbn-products} \cite{AmazonProducts} datasets from the Open Graph Benchmark (OGB) \cite{OGB}, which are two of the most widely used node property prediction benchmarking datasets with contrasting size and density. 
\textit{Ogbn-arxiv} is a citation network between arXiv papers, where the nodes are papers and the edges indicate when one paper cites another. The nodes' text attributes are papers' titles and abstracts. The node classes to be predicted are the subject areas of the papers. \textit{Ogbn-products} is an Amazon product network, where the nodes are products and the edges indicate when two products are purchased together. The nodes' text attributes are products' titles and descriptions. The node classes to be predicted are the products' categories. The statistics of the datasets are shown in Table \ref{tab:data_stats}.

\begin{table}
\centering
\begin{tabular}{lrrc}
\hline
              & \#Nodes & \#Edges & Split ratio (\%) * \\ \hline
ogbn-arxiv    & 0.17M   & 1.17M   & 54/18/28           \\
ogbn-products & 2.44M   & 61.86M  & 8/2/90             \\ \hline
\multicolumn{4}{l}{* train/validation/test split}     
\end{tabular}
\caption{Open Graph Benchmark (OGB) data statistics.}
\label{tab:data_stats}
\end{table}

E2EG's hyperparameters are mostly similar to GIANT. We tune some of them to maximize the validation accuracy. These include maximum learning rate with values $[3 \times 10^{-4}, 1 \times 10^{-4}, 6 \times 10^{-5}, 1 \times 10^{-5}]$, dropout probability with values $[0.01, 0.05, 0.1, 0.2]$, number of layers in the main node classification head with values $[1, 2, 3]$. The final hyperparameters are as follows: linearly decay learning rate with maximum value of $6\times 10^{-5}$, batch size of 32, dropout probability of 0.1, number of fine-tuning rounds for neighborhood prediction (i.e., HLT depth) of 4 for \textit{ogbn-arxiv} and 2 for \textit{ogbn-products}, weighted-squared-hinge loss for neighborhood prediction and the cross-entropy loss for node classification. We use half of the validation and test data for neighborhood prediction training in the transductive setting for \textit{ogbn-products}, since the portion of test data here is notably large. The main node classification head is a single linear projection layer that maps text's hidden dimension to the number of classes. Each multi-task fine-tuning round is trained for 1 epoch. The additional round is trained with early stopping, i.e., trained until the validation accuracy on the main task stops increasing. For \textit{ogbn-arxiv}, we exclude the main task in the first 2 fine-tuning rounds (also determined by hyperparameter tuning) to avoid overfitting.

We compare E2EG to some baselines:
\begin{itemize}
    \item A baseline that \textit{only uses graph topology} for node classification. We choose GraphSAGE with nodes' degrees input, omitting the use of the nodes’ text attributes.
    \item A baseline that \textit{only uses the nodes’ text attributes} for node classification. We choose the BERT model for text classification, taking only the node's raw text as input, thus omitting the use of graph topology.
    \item Architectures that \textit{make use of both} nodes’ text attributes and graph topology, but \textit{not end-to-end}: 1) GraphSAGE, using OGB's graph-agnostic text embeddings and 2) GIANT+MLP, using graph-dependent text embeddings.
\end{itemize}

We do not include complex ensemble pipelines (e.g., GIANT+GraphSAGE or GIANT+DRGAT+KD) as baselines, since these pipelines are more costly to train with a high number of parameters, thus incomparable to our proposed end-to-end model. Among the baselines, we especially do a comprehensive comparison between our E2EG model and the GIANT+MLP baseline, since we are extending from GIANT+MLP. We compare E2EG with GIANT+MLP when using different transformer versions: the original BERT encoder and the more lightweight DistilBERT and Distil-SentenceBERT encoders. We also compare them in the inductive setting, where the nodes in the test and validation set are not used to learn neighborhood prediction. 

Although E2EG is designed to be a stand-alone node classification model, we additionally experiment with using E2EG to generate text embeddings for other downstream classifiers in the same way as GIANT. We conduct this experiment by replacing the embeddings generated by GIANT's text encoder with the embeddings generated by E2EG's text encoder in one of the top node classification pipelines on the OGB leaderboards for \textit{ogbn-arxiv} (i.e., GIANT+DRGAT+KD) and \textit{ogbn-products} (i.e., GIANT+SAGN+MCR+C\&S).

To study E2EG's sources of performance gain, we perform qualitative analysis by investigating the samples predicted correctly by E2EG, but incorrectly by BERT or GIANT+MLP. We additionally investigate the samples predicted incorrectly by E2EG. For each model, we consider different versions trained with different random seeds, and look into the samples that are predicted correctly or incorrectly by all versions. We look at the text of node samples and their two-hop neighborhood as an indication of graph topology.

The experiments are done on machines with 2 NVIDIA GeForce GTX 1080 Ti GPUs with 11GB memory each and 12 CPUs with 128GB RAM in total. We repeat each experiment over 10 random seeds from 0 to 9 and report the average and standard deviation of node classification accuracies.

\section{Results} \label{sec:Results}

\subsection{Effect of proposed training strategies}
\begin{table}
\centering
\begin{tabular}{lccc}
\hline
                                                                              & \begin{tabular}[c]{@{}c@{}}Best validation \\ accuracy (\%)\end{tabular} & \begin{tabular}[c]{@{}c@{}}Best \\ round *\end{tabular} & \begin{tabular}[c]{@{}c@{}}Test \\ accuracy (\%)\end{tabular} \\ \hline
Base model                                                                    & 74.41 ± 0.09                                                             & 2/4                                                     & 73.06 ± 0.14                                                  \\
With delay & \textbf{74.81 ± 0.16}                                                    & \textbf{4/4}                                            & \textbf{73.60 ± 0.16}                                         \\ \hline
\multicolumn{4}{l}{\begin{tabular}[c]{@{}l@{}}*: \textit{best\_round / total\_number\_of\_rounds}, where \textit{best\_round} \\is the round with the highest validation accuracy\end{tabular}}                                                                                        
\end{tabular}
\caption{Effect of E2EG's main-task-delay strategy on 
\textit{ogbn-arxiv}.}
\label{tab:tackle_overfitting}
\end{table}

The effect of delaying the main task for E2EG is shown in Table \ref{tab:tackle_overfitting}. We observe that the overfitting issue, i.e., the best validation score obtained at an early fine-tuning round, only occurs for \textit{ogbn-arxiv}. Therefore, we only apply this method on \textit{ogbn-arxiv}. Observe that the base model's best validation accuracy is obtained early at the second round (out of 4 rounds in total). Excluding the main loss in the early rounds helps avoid overfitting, as the best validation score increases by 0.4\% and is obtained at the last fine-tuning round. Using this method, the model's test accuracy is improved by 0.54\%.

The effect of including an additional main-task-only fine-tuning round is shown in Table \ref{tab:additionalRounds}. As can be seen, the additional rounds lead to test accuracy improvements of +0.56\% on \textit{ogbn-arxiv} and +0.71\% on \textit{ogbn-products}.

\begin{table}
\centering
\begin{tabular}{llcc}
\hline
                                                                          &          & \begin{tabular}[c]{@{}c@{}}Validation \\ accuracy\end{tabular} & \begin{tabular}[c]{@{}c@{}}Test \\ accuracy\end{tabular} \\ \hline
\multirow{2}{*}{\begin{tabular}[c]{@{}l@{}}ogbn-arxiv\\~\end{tabular}}    & Before * & 74.41 ± 0.09                                                   & 73.06 ± 0.14                                             \\
                                                                          & After *  & \textbf{74.87 ± 0.11}                                          & \textbf{73.62 ± 0.14}                                    \\ \hline
\multirow{2}{*}{\begin{tabular}[c]{@{}l@{}}ogbn-products\\~\end{tabular}} & Before * & 92.24 ± 0.18                                                   & 80.27 ± 0.32                                             \\
                                                                          & After  * & \textbf{92.34 ± 0.09}                                          & \textbf{80.98 ± 0.40}                                    \\ \hline
\multicolumn{4}{l}{\begin{tabular}[c]{@{}l@{}}*: performance before and after the additional rounds\end{tabular}}                                                                                
\end{tabular}
\caption{Effect of E2EG's additional fine-tuning rounds.}
\label{tab:additionalRounds}
\end{table}

The effect of transductive/inductive data utilization and lightweight encoders are reported along with comparisons to the GIANT+MLP baseline in the following sections.

\subsection{Baselines performance}

\begin{table*}[t]
\centering
\begin{tabular}{@{}rlll@{}}
\hline
\multicolumn{1}{l}{No.}                                                                                                                                    & \multicolumn{1}{c}{Model}                                                                                                                                                                        & \multicolumn{2}{l}{\hspace{2cm}Accuracy (\%)}                                                                                                                                                                                                                                                                          \\
\multicolumn{1}{l}{}                                                                                                                                       &                                                                                                                                                                                                  & ogbn-arxiv                                                                                                                                             & ogbn-products                                                                                                                                          \\ \hline
1                                                                                                                                                          & BERT text classifier                                                                                                                                                           & 69.66 ± 0.50                                                                                                                                           & 76.04 ± 0.72                                                                                                                                           \\
2                                                                                                                                                          & GraphSAGE with node degree                                                                                                                                    & 39.57 ± 0.59                                                                                                                                           & 39.00 ± 0.70                                                                                                                                           \\
3                                                                                                                                                          & GraphSAGE with nodes' graph-agnostic embedded text                                                                                                                                                           & 71.49 ± 0.27 *                                                                                                                                         & 78.50 ± 0.14 *                                                                                                                                         \\
4                                                                                                                                                          & GIANT+MLP                                                                                                                                                                                        & 73.06 ± 0.11 *                                                                                                                                         & 80.49 ± 0.28 *                                                                                                                                         \\ \hline
5                                                                                                                                                          & GIANT+MLP - BERT $^a$                                                                                                                                                                            & 73.06 ± 0.11 *                                                                                                                                         & 80.49 ± 0.28 *                                                                                                                                         \\
6                                                                                                                                                          & GIANT+MLP - DistilBERT $^b$                                                                                                                                                                      & 70.99 ± 0.16                                                                                                                                           & 79.49 ± 0.25                                                                                                                                           \\
7                                                                                                                                                          & GIANT+MLP - Distil-SentenceBERT $^c$                                                                                                                                                      & 72.50 ± 0.25                                                                                                                                           & 79.46 ± 0.23                                                                                                                                           \\
8                                                                                                                                                          & GIANT+MLP - inductive                                                                                                                                                                        & 70.46 ± 0.22                                                                                                                                           & 77.52 ± 0.25                                                                                                                                           \\ \hline
9                                                                                                                                                         & E2EG - BERT $^a$                                                                                                                                                                                 & 73.12 ± 0.20 (+0.06)                                                                                                                                   & \textbf{81.11 ± 0.37} (+0.62)                                                                                                                                   \\
10                                                                                                                                                         & E2EG - DistilBERT $^b$                                                                                                                                                                           & 72.97 ± 0.19 (+1.98)                                                                                                                                   & 80.98 ± 0.40 (+1.49)                                                                                                                                   \\
11                                                                                                                                                         & E2EG - Distil-SentenceBERT $^c$                                                                                                                                                           & \textbf{73.62 ± 0.14} (+1.12)                                                                                                                                   & 80.88 ± 0.40 (+1.42)                                                                                                                                   \\
12                                                                                                                                                         & E2EG - inductive                                                                                                                                                                            & 72.69 ± 0.17 (+2.23)                                                                                                                                   & 78.57 ± 0.26 (+1.05)                                                                                                                                   \\ \hline
\multicolumn{4}{l}{\begin{tabular}[c]{@{}l@{}}
*: Results from OGB leaderboard.\hspace{2cm} $^{a,b,c}$: Models with different types of the transformer component.\\ Numbers in parentheses in rows 9-12 for E2EG are comparisons to the corresponding settings in rows 5-8 for GIANT+MLP.\end{tabular}}
\end{tabular}
\caption{Node classification performance of the proposed E2EG model in comparison with the baselines.}
\label{tab:big_table}
\end{table*}

The performance of the baselines is shown in Row 1-4 in Table \ref{tab:big_table}. GraphSAGE with node degree, using only graph topology, has the worst performance, at under 40\% for both datasets. On the other hand, GraphSAGE with graph-agnostic embedded text has notably better performance, almost double the accuracy compared to GraphSAGE with node degree. The BERT text classifier also performs well without making use of graph topology, achieving 69.66\% accuracy on \textit{ogbn-arxiv} and 76.04\% on \textit{ogbn-products}. GIANT+MLP has the best performance with 73.06\% accuracy on \textit{ogbn-arxiv} and 80.49\% accuracy on \textit{ogbn-products}.

\subsection{E2EG versus the baselines}
The performance of E2EG in comparison with the baselines is shown in Table \ref{tab:big_table}. Numbers in parentheses in rows 9-12 for E2EG are comparisons to the corresponding settings in rows 5-8 for GIANT+MLP. Overall, in the default transductive setting, the best E2EG model outperforms all selected baselines. It outperforms GraphSAGE with graph-agnostic text embeddings by +2.13\% on \textit{ogbn-arxiv} and +2.61\% on \textit{ogbn-products}. E2EG also outperforms the BERT text classifier by +3.96\% on \textit{ogbn-arxiv} and +5.07\% on \textit{ogbn-products}. Most importantly, E2EG slightly outperforms the strong GIANT+MLP baseline by over +0.5\% for both datasets.

The effect of using a more lightweight text encoder for E2EG is shown in Row 9-11, in comparison with GIANT+MLP in Row 5-7 in Table \ref{tab:big_table}. When BERT is used, the performance of E2EG is similar to GIANT+MLP. Using DistilBERT instead of BERT worsens GIANT+MLP's accuracy by up to -2.07\%. On the other hand, E2EG maintains its performance when using DistilBERT, with a negligible decrease of at most -0.15\% in accuracy. Using Distil-SentenceBERT leads to accuracy improvement for E2EG on \textit{ogbn-arxiv} by +0.65\%, while having similar performance on \textit{ogbn-products} as DistilBERT. The use of distilled transformers results in a reduction in the number of parameters, by -40\% with DistilBERT and -25\% with Distil-SentenceBERT, leading to better runtime. The details on models' numbers of parameters and runtime when trained on the same machine are in Table \ref{tab:runtime}. 

\begin{table}
\centering
\begin{tabular}{llrr}
\hline
                                                                          &                & \multicolumn{1}{c}{\begin{tabular}[c]{@{}c@{}}\#Params\end{tabular}} & \multicolumn{1}{c}{\begin{tabular}[c]{@{}c@{}}Train time (h)\end{tabular}} \\ \hline
\multirow{2}{*}{\begin{tabular}[c]{@{}l@{}}ogbn-arxiv\\~\end{tabular}}    & GIANT+MLP $^a$ & 111M                                                                         & 3.27 ± 0.07                                                                         \\
                                                                          & E2EG $^b$      & \textbf{84M}                                                                 & \textbf{2.94 ± 0.23}                                                                \\
\multirow{2}{*}{\begin{tabular}[c]{@{}l@{}}ogbn-products\\~\end{tabular}} & GIANT+MLP $^a$ & 110M                                                                         & 16.85 ± 0.77                                                                        \\
                                                                          & E2EG $^c$      & \textbf{67M}                                                                 & \textbf{9.89 ± 1.21}                                                                \\ \hline
\multicolumn{4}{l}{\begin{tabular}[c]{@{}l@{}}$^a$: Use BERT text encoder\\ $^b$: Use Distil-SentenceBERT text encoder \\ $^c$: Use DistilBERT text encoder\end{tabular}}                                                                               
\end{tabular}
\caption{E2EG versus GIANT+MLP on computational cost.}
\label{tab:runtime}
\end{table}

The result of using E2EG in an inductive setting, in comparison with GIANT+MLP, is shown in Row 8 and Row 12 in Table \ref{tab:big_table}. Our model is better than GIANT+MLP by +2.23\% on \textit{ogbn-arxiv} and +1.05\% on \textit{ogbn-products}.

We additionally show the result of using E2EG only to generate text embeddings, in comparison with GIANT, in Table \ref{tab:ensemble}. The generated embeddings are used in ensemble pipelines that chain multiple models together for node classification. As can be seen, the performance of the pipelines using E2EG's embeddings falls short of the ones using GIANT's embeddings by -0.72\% on \textit{ogbn-arxiv} and -2.16\% on \textit{ogbn-products}.

\begin{table}
\centering
\begin{tabular}{lll}
\hline
                                                                          & Ensemble pipeline         & Accuracy (\%) \\                                  \hline
\multirow{2}{*}{\begin{tabular}[c]{@{}l@{}}ogbn-arxiv\\~\end{tabular}}    & GIANT+DRGAT+KD      & \textbf{76.33 ± 0.08}                                                                              \\
                                                                          & E2EG+DRGAT+KD             & 75.61 ± 0.56                                                                                       \\ \hline
\multirow{2}{*}{\begin{tabular}[c]{@{}l@{}}ogbn-products\\~\end{tabular}} & GIANT+SAGN+MCR+C\&S & \textbf{86.73 ± 0.08}                                                                              \\
                                                                          & E2EG+SAGN+MCR+C\&S        & 84.57 ± 0.06                                                                                       \\ \hline
                                                                                                                                  
\end{tabular}
\caption{E2EG versus GIANT in ensemble pipelines.}
\label{tab:ensemble}
\end{table}

\subsection{Qualitative analysis}

\begin{figure}[h]
  \centering
  \subfloat[Correctly predicted by E2EG, incorrect by BERT]{
  \includegraphics[width=0.95\columnwidth]{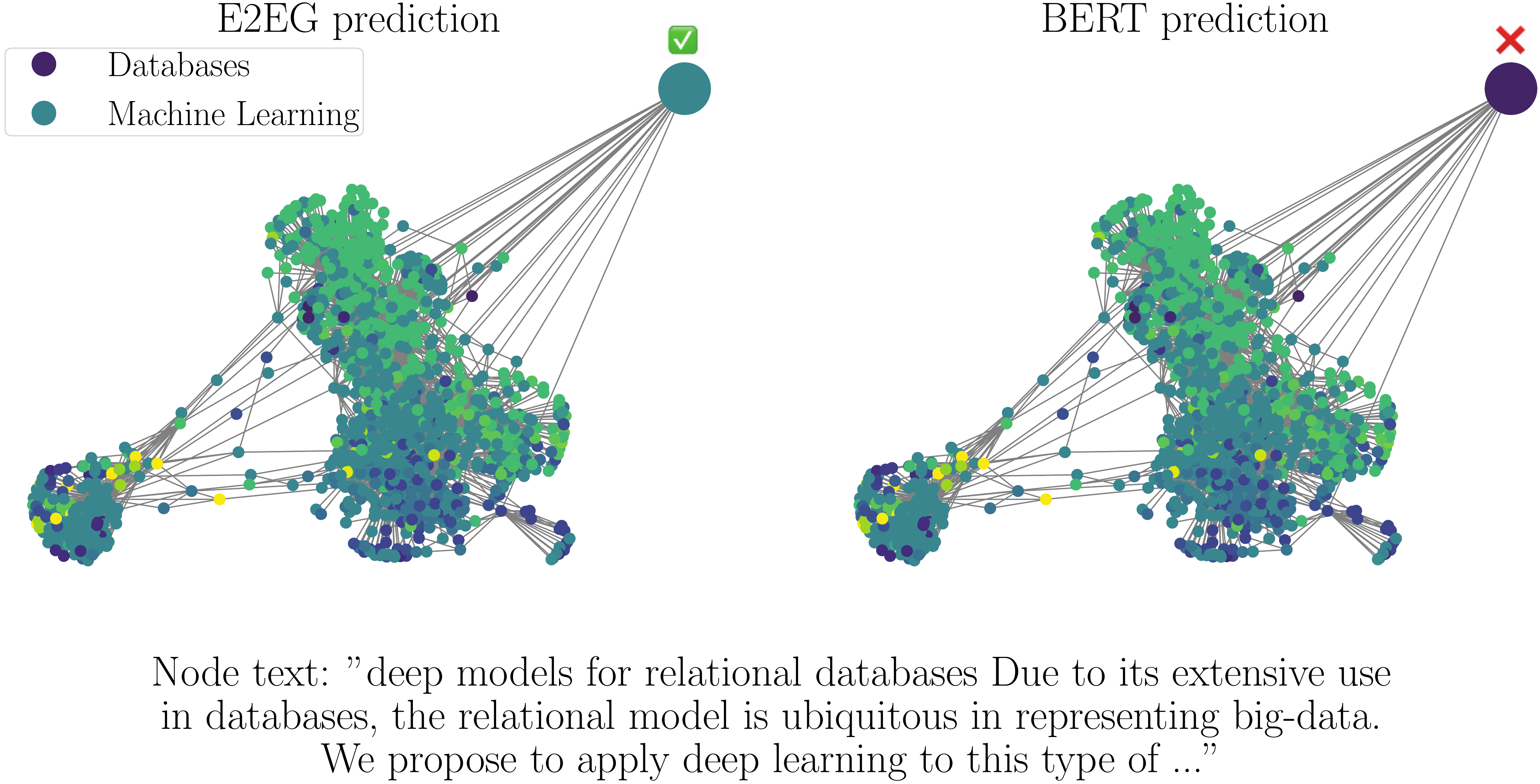}
}\\
  \subfloat[Correctly predicted by E2EG, incorrect by GIANT+MLP]{
  \includegraphics[width=0.95\columnwidth]{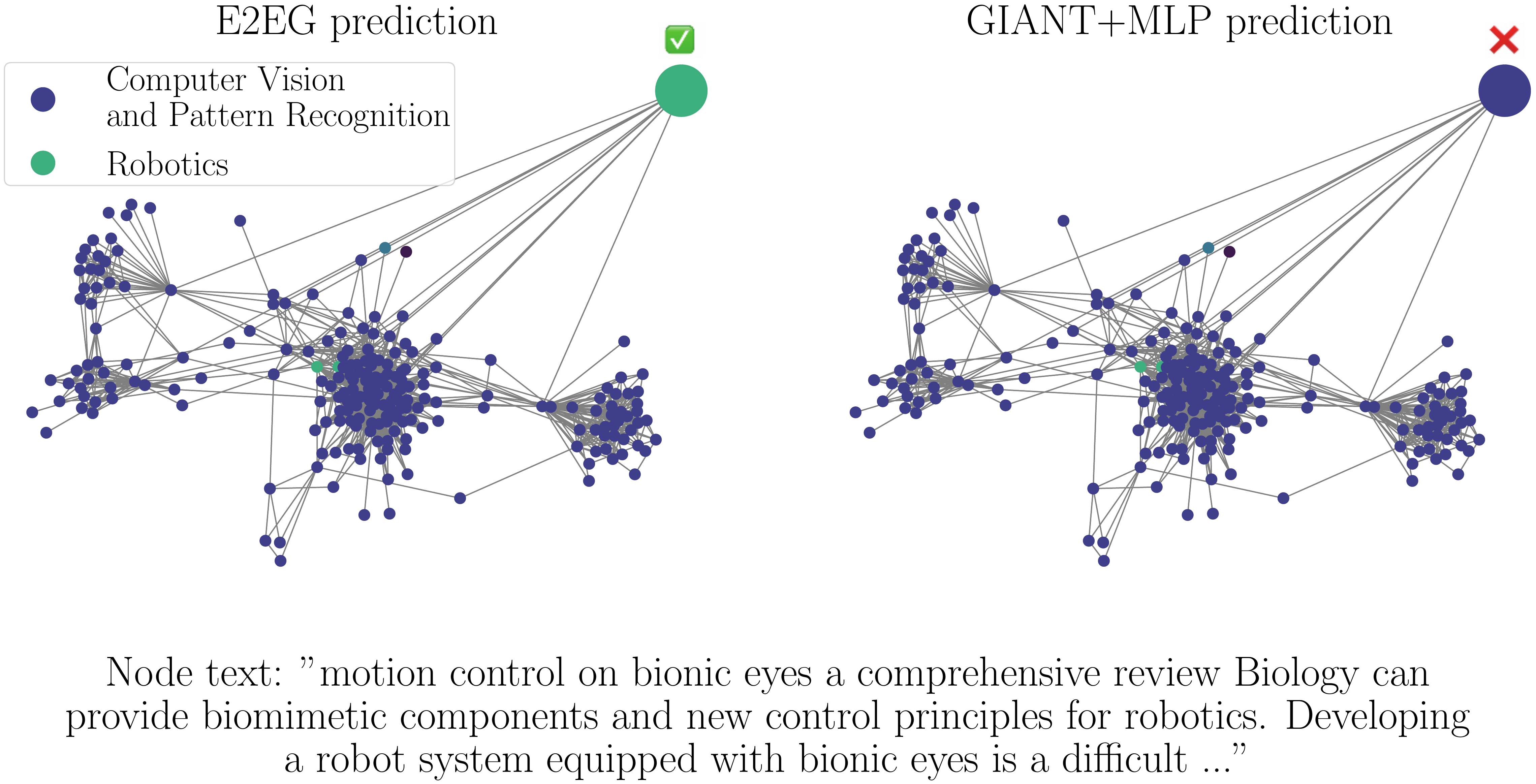}
}\\
  \subfloat[Sample node incorrectly predicted by E2EG]{
  \includegraphics[width=0.95\columnwidth]{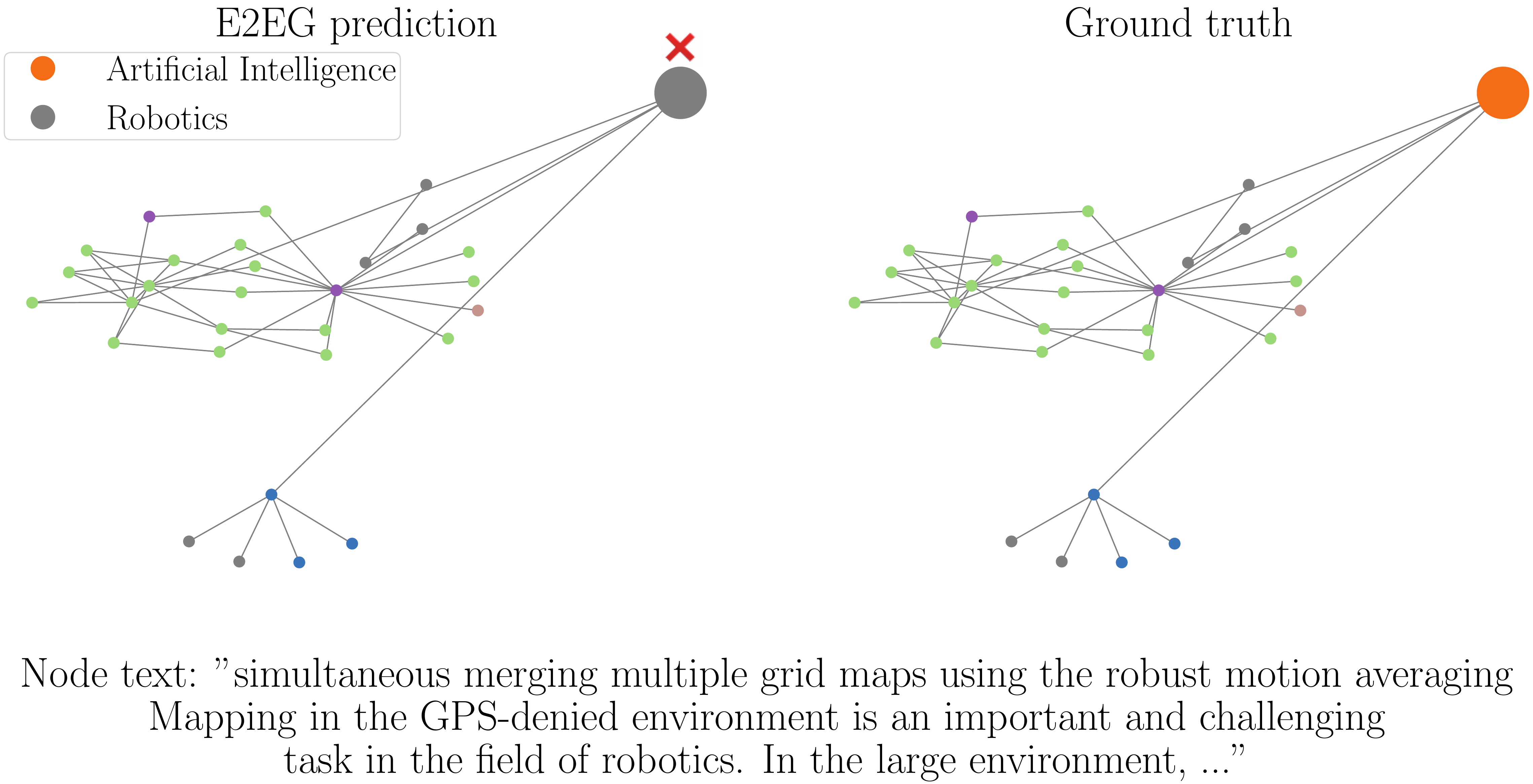}
}

  \caption{Samples from \textit{ogbn-arxiv}. The predicted node is the big circle at the top right. The other circles represent the two-hop neighborhood of the predicted node. The big node’s color represents the predicted class, the other nodes' colors represent the actual classes. The text at the bottom is the predicted node's raw text.}
  \label{fig:qualitative_arxiv_samples}
\end{figure}

We observe that node samples predicted correctly by E2EG but incorrectly by BERT have informative topological information, which E2EG succeeded to exploit in contrast to BERT. An example is shown in Figure \ref{fig:qualitative_arxiv_samples}a. E2EG predicts the node topic correctly as \textit{Machine Learning}, while BERT predicts the node topic incorrectly as \textit{Databases}. From the node's raw text, it is difficult to tell whether the class is \textit{Machine Learning} or \textit{Databases}, as it is about \textit{"Deep models for relational databases."}. In contrast, the two-hop neighborhood is more informative, as the majority of the neighborhood has the same \textit{Machine Learning} class as the predicted node.

We observe that node samples predicted correctly by E2EG but incorrectly by GIANT+MLP have informative text attributes, which E2EG succeeded to exploit in contrast to GIANT+MLP. An example is shown in Figure \ref{fig:qualitative_arxiv_samples}b. E2EG predicts the node topic correctly as \textit{Robotics}, while GIANT+MLP predicts it incorrectly as \textit{Computer Vision and Pattern Recognition}. From the node's raw text, it is clear that the class is \textit{Robotics}, since the text mentions "motion control" and "robot system". In contrast, the two-hop neighborhood is quite misleading, as the majority of the neighborhood is from the \textit{Computer Vision and Pattern Recognition} class.

We observe that node samples predicted incorrectly by E2EG are misleading in both modalities: text and topological information. An example is shown in Figure \ref{fig:qualitative_arxiv_samples}c. E2EG predicts the node topic incorrectly as \textit{Robotics}, while the correct class is \textit{Artificial Intelligence}. The node's raw text is quite misleading, as it is about an artificial intelligence task "in the field of robotics". The neighborhood is also not very informative, since none of the neighbors has the same \textit{Artificial Intelligence} class as the considered node sample.

\section{Discussion} \label{sec:Discussion}
We have observed that our proposed training strategy to delay learning the main task helps improve E2EG's performance on a smaller dataset that suffers from overfitting, i.e., \textit{ogbn-arxiv}. Including an additional round learning only the main task, has shown to improve E2EG's performance across both datasets.

The baseline results indicate the crucial role played by text attributes in node classification. GraphSAGE performs notably worse with node degrees features in comparison to text embedding features. The BERT text classifier has decent performance without making use of graph topology, suggesting that for node classification on \textit{ogbn-arxiv} and \textit{ogbn-products}, the nodes' text attributes are the more informative modality compared to graph topology. It also shows the effectiveness of embedding nodes' raw text directly for node classification, which is one of the motivations for our end-to-end model.

It has been shown that E2EG outperforms some baselines on node classification. E2EG performs better than the BERT text classifier. This shows that E2EG's effectiveness comes from the multi-task setting with the auxiliary neighborhood prediction task, and not only from embedding the raw text directly for node classification. E2EG also outperforms GraphSAGE without the need to use computationally-expensive message passing. Most importantly, E2EG slightly outperforms GIANT+MLP, showing the effectiveness of the end-to-end fashion. E2EG also outperforms GIANT+MLP in the inductive setting. This makes E2EG applicable in real-world scenarios, where graph data such as social networks or citation networks constantly changes with new nodes and edges.

Our experimental results indicate that E2EG is expressive enough to utilize a more lightweight text encoder for a negligible loss in accuracy. When BERT is used, the performance of E2EG is similar to GIANT+MLP, since BERT is a complex text encoder that probably captured most of the information needed from the raw text during the embedding process. In contrast, when using DistilBERT or Distil-SentenceBERT, GIANT+MLP suffers increased accuracy loss, possibly due to increased information loss during the text-embedding stage when using a more lightweight encoder. Our proposed E2EG model was able to maintain its performance, strengthening our hypothesis that the end-to-end process with an expressive learning objective allows for the use of a more lightweight text encoder. E2EG's performance on \textit{ogbn-arxiv} is further improved by using Distil-SentenceBERT instead of DistilBERT. This strengthens the hypothesis that sentence-level pretraining is more suitable. For \textit{ogbn-products}, the effect of sentence-level pretraining is not notable, potentially due to the larger amount of data and less complex sentences compared to the academic \textit{ogbn-arxiv} dataset.

While E2EG outperforms the GIANT+MLP baseline in a standalone fashion, it falls short behind GIANT when being used to generate text embeddings for downstream classifiers. This can be seen as a limitation of E2EG's applicability. However, note that ensemble pipelines involve multiple chained models. Meanwhile, E2EG is designed to be used in an end-to-end manner. This would make E2EG more suitable for real-world scenarios, since the model would be compact and stand-alone, while maintaining decent performance.

The qualitative analysis provides insights into E2EG's performance gain. We have observed that E2EG's correct predictions over BERT possibly come from the exploitation of graph topology. This is an indication of the effectiveness of the multi-task setting with an auxiliary neighborhood prediction task for utilizing topological information. On the other hand, it has been seen that E2EG's correct predictions over GIANT+MLP possibly come from the better exploitation of nodes' raw text attributes. This is an indication of the effectiveness of the end-to-end fashion: it reduces information loss by embedding the raw text directly for node classification. Overall, the observed patterns indicate that E2EG better exploits both nodes' raw text and graph topology.

\section{Conclusion} \label{sec:Conclusion}
We introduced an end-to-end node classification model termed E2EG. E2EG simultaneously learns two tasks: the neighborhood prediction task to encode graph topology, and the main node classification task. E2EG offers benefits compared to previous two-stage approaches: it allows for lightweight text encoder usage and potentially reduces information loss. Our experiments indicate accuracy improvements of +0.5\% compared to GIANT+MLP in the default transductive setting on \textit{ogbn-arxiv} and \textit{ogbn-products}, while using a faster, distilled transformer component with 25\% - 40\% fewer parameters. E2EG is also applicable in an inductive setting, with a gain of up to +2.23\% in accuracy compared to GIANT+MLP. E2EG is more suitable to be used in a stand-alone manner rather than in a complicated ensemble pipeline. This makes E2EG well-suited in real-world scenarios where a compact model is preferred. Our qualitative analysis indicates that the performance gain of E2EG is due to the multi-task and the end-to-end setting that leads to better exploitation of the graph topology and nodes' raw text.

\bibliography{references}
\bibliographystyle{IEEEtran}

\end{document}